%% file: aaai24.tex
\documentclass[letterpaper]{article} % DO NOT CHANGE THIS
\pdfoutput=1
\usepackage{aaai24}  % DO NOT CHANGE THIS
\usepackage{times}  % DO NOT CHANGE THIS
\usepackage{helvet}  % DO NOT CHANGE THIS
\usepackage{courier}  % DO NOT CHANGE THIS
\usepackage[hyphens]{url}  % DO NOT CHANGE THIS
\usepackage{graphicx} % DO NOT CHANGE THIS
\usepackage{natbib}  % DO NOT CHANGE THIS AND DO NOT ADD ANY OPTIONS TO IT
\usepackage{caption} % DO NOT CHANGE THIS AND DO NOT ADD ANY OPTIONS TO IT
\usepackage{algorithm}
\usepackage{algorithmic}
\usepackage{graphicx}
\usepackage{amsmath}
\usepackage{amssymb}
\usepackage{booktabs}
\usepackage{multirow}
\usepackage{amsmath,amsfonts}
\usepackage{algorithmic}
\usepackage{algorithm}
\usepackage{array}
\usepackage[caption=false,font=normalsize,labelfont=sf,textfont=sf]{subfig}
\usepackage{textcomp}
\usepackage{url}
\usepackage{verbatim}
\usepackage{graphicx}
\usepackage{cite}
\usepackage{graphicx}
\usepackage{booktabs}
\usepackage{times}
\usepackage{graphicx}
\usepackage{epstopdf}
\usepackage{verbatim}
\usepackage{epsfig}
\usepackage{amsmath}
\usepackage{amssymb}
\usepackage{color}
\usepackage{booktabs}
\usepackage{xcolor}
\usepackage{color, colortbl}
\usepackage{makecell}
\usepackage{amsmath,amsfonts}
\usepackage{algorithmic}
\usepackage{algorithm}
\usepackage{array}
\usepackage[caption=false,font=normalsize,labelfont=sf,textfont=sf]{subfig}
\usepackage{textcomp}
\usepackage{url}
\usepackage{verbatim}
\usepackage{graphicx}
\usepackage{cite}
\usepackage{times}
\usepackage{graphicx}
\usepackage{epstopdf}
\usepackage{verbatim}
\usepackage{epsfig}
\usepackage{amsmath}
\usepackage{amssymb}
\usepackage{color}
\usepackage{multirow}
\usepackage{makecell}
\usepackage{array}
\usepackage{booktabs}
\usepackage{xcolor}
\usepackage{color, colortbl}
\usepackage[algo2e]{algorithm2e}

\usepackage{multirow}
\usepackage{xcolor}
\usepackage{color, colortbl}

\usepackage{fancyhdr}
\usepackage{lipsum} % for dummy text
\usepackage{newfloat}
\usepackage{listings}
\DeclareCaptionStyle{ruled}{labelfont=normalfont,labelsep=colon,strut=off} % DO NOT CHANGE THIS
\floatstyle{ruled}
\newfloat{listing}{tb}{lst}{}
\floatname{listing}{Listing}
\nocopyright
\title{TS-SatMVSNet: Slope Aware Height Estimation for Large-Scale Earth Terrain Multi-view Stereo}
\author{
    Song Zhang \textsuperscript{\rm 1 \rm 2} 
    Zhiwei Wei \thanks{Corresponding author} \textsuperscript{\rm 1},
    Wenjia Xu \textsuperscript{\rm 3},
    Lili Zhang \textsuperscript{\rm 1},
    Yang Wang \textsuperscript{\rm 1},
    Jinming Zhang \textsuperscript{\rm 1},
    Junyi Liu \textsuperscript{\rm 1}
}
\affiliations{
    \textsuperscript{\rm 1}Aerospace Information Research Institute, Chinese Academy of Sciences. 
    \textsuperscript{\rm 2}University of Chinese Academy of Sciences.
    \textsuperscript{\rm 3}Beijing University of Posts and Telecommunications.\\
}

\fancypagestyle{firstpage}{
  \fancyhf{} % clear all header and footer fields
  \fancyfoot[L]{\small{This work has been submitted to the IEEE for possible publication. Copyright may be transferred without notice, after which this version may no longer be accessible.}}
}

\usepackage{bibentry}
\begin{document}

\maketitle

% Add the custom footer for the first page
\thispagestyle{firstpage}

\begin{abstract}
% Large-scale reconstruction of the Earth’s surface, using remote sensing imagery, is a crucial research problem in the field of remote sensing. Given that the continental regions of the Earth are primarily composed of terrains, the height estimation of terrains becomes an essential task in 3D reconstruction. Recently, learning-based multi-view stereo (MVS) methods have demonstrated promising performance in the field of remote sensing height estimation. Despite the efforts made by previous MVS methods to modify the general learning-based MVS framework, the characteristics of terrains have not been fully taken into account, leading to suboptimal accuracy. As is well known, the earth’s surface is generally undulating but does not undergo drastic changes, therefore, it can be measured by the slope. In this work, we propose an end-to-end slope aware terrain height estimation network for large-scale remote sensing reconstruction, called TE-SatMVSNet. 
3D terrain reconstruction with remote sensing imagery achieves cost-effective and large-scale earth observation and is crucial for safeguarding natural disasters, monitoring ecological changes, and preserving the environment.
% 3D terrain reconstruction using remote sensing imagery achieves cost-effective and large-scale means of observing the earth's surface and is crucial for safeguarding natural disasters, monitoring ecological changes, and preserving the environment. 
Recently, learning-based multi-view stereo~(MVS) methods have shown promise in this task. However, these methods simply modify the general learning-based MVS framework for height estimation, which overlooks the terrain characteristics and results in insufficient accuracy. Considering that the Earth's surface generally undulates with no drastic changes and can be measured by slope, integrating slope considerations into MVS frameworks could enhance the accuracy of terrain reconstructions.
To this end, we propose an end-to-end slope-aware height estimation network named TS-SatMVSNet for large-scale remote sensing terrain reconstruction.
% From a macro perspective, the slope trend of terrain in remote sensing will be reflected as changes in slope direction, and the slope can be measured by the difference in height. Therefore, we first proposed a strategy to calculate the slope map based on the height map to measure the undulating trend of the terrain. Based on the above strategy, we proposed a slope constraint to optimize the results of height estimation. In addition, by using the height difference, we designed a slope-guided height interval adaptive division module to achieve more refined height estimation. From a micro perspective, the terrain is composed of countless tiny planes. Based on the above assumption, we designed a height correction module to correct the wrong height values using Gaussian smoothing filtering to achieve accurate height estimation.
To effectively obtain the slope representation, drawing from mathematical gradient concepts, we innovatively proposed a height-based slope calculation strategy to first calculate a slope map from a height map to measure the terrain undulation. To fully integrate slope information into the MVS pipeline, we separately design two slope-guided modules to enhance reconstruction outcomes at both micro and macro levels. Specifically, at the micro level, we designed a slope-guided interval partition module for refined height estimation using slope values. At the macro level, a height correction module is proposed, using a learnable Gaussian smoothing operator to amend the inaccurate height values. Additionally, to enhance the efficacy of height estimation, we proposed a slope direction loss for implicitly optimizing height estimation results. Extensive experiments on the WHU-TLC dataset and MVS3D dataset show that our proposed method achieves state-of-the-art performance and demonstrates competitive generalization ability compared to all listed methods. % Specifically, xxx.
% Our code will be available at XXX.
\end{abstract}

\begin{figure}[!t]
    \centering 
    \includegraphics[width=0.95\linewidth]{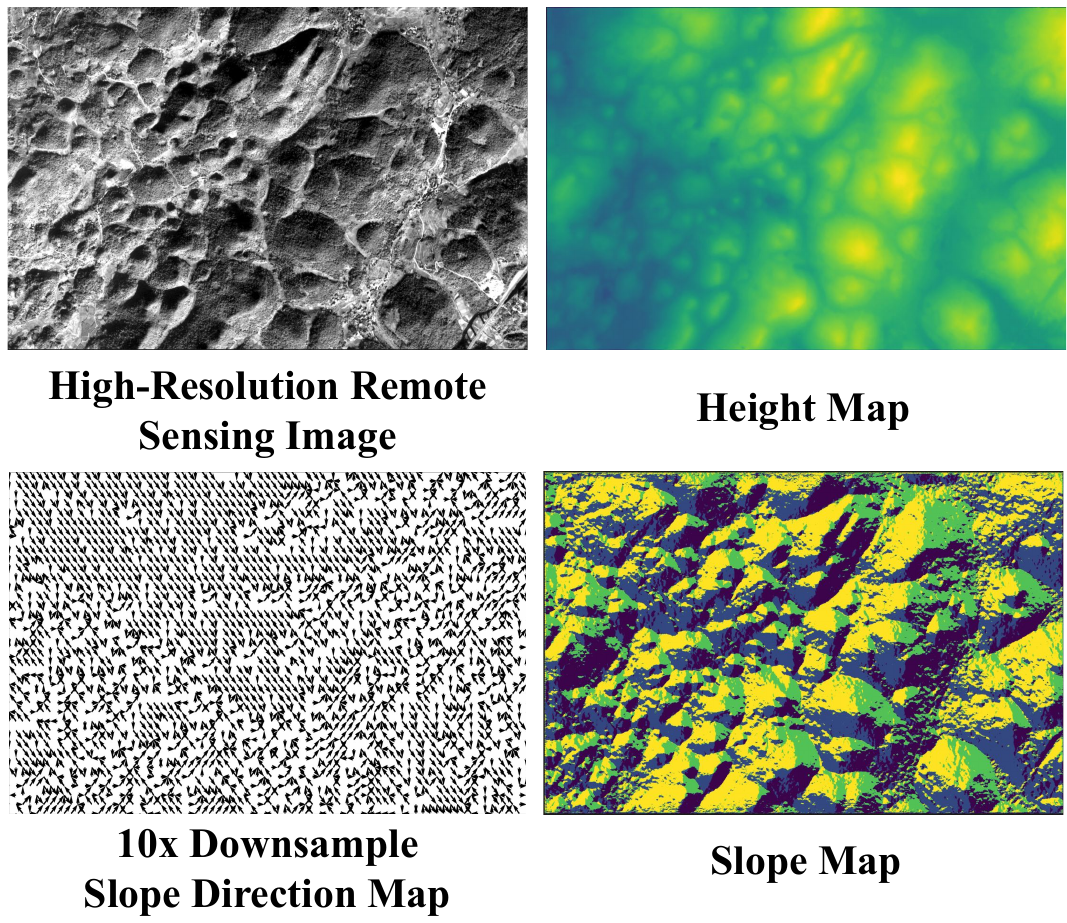}\vspace{-3mm}
    \caption{Illustration of High-Resolution Remote Sensing Image, Height Map, 10x Downsample Slope Direction Map and Slop Map based on WHU-TLC dataset \cite{satmvs}. We use 10x downsampling for the visualization of the slope direction map.}
    \label{fig:figure_1} 
    \vspace{-3mm}
\end{figure}

\section{Introduction}
Large-scale reconstruction of the earth’s surface provides valuable insights into the earth's features and is particularly important for applications such as monitoring ecological changes \cite{monitoring}, detecting geographic information \cite{geographical}, and providing early warnings for natural disasters \cite{prendes2014new}. Remote sensing imagery facilitates cost-effective and extensive earth observation and has become a crucial data source for the task \cite{whitaker2002reconstruction}. Considering that the Earth’s continental regions are primarily composed of terrains, accurate terrain height estimation using remote sensing imagery is crucial for 3D reconstruction of the Earth’s surface and numerous algorithms have been developed for this task with rational polynomial camera (RPC) parameters\cite{automatic, bundle, rpc}. 
% However, these classical open-source algorithms for large-scale 3D terrain reconstruction may not always be effective. This is due to traditional feature extraction operators \cite{sift} may not be able to fully extract complex terrain surface features, which may lead to error height estimation and subsequently cause reconstruction holes.
However, these classical open-source algorithms for large-scale 3D terrain reconstruction usually use traditional feature extraction operators \cite{sift} that are not able to fully extract complex terrain surface features, which may lead to error height estimation and subsequently cause reconstruction holes.

% 尽管一些经典的开源算法可以结合有理多项式相机（RPC）参数用于从卫星立体对中提取三维信息，但是对于大规模三维重建方法仍然是十分低效的，同时由于传统的特征提取算法可能对复杂的地表特征无法充分的提取，可能会导致错误的高度估计，进而导致重建空洞的问题。

Recently, the advancement of deep learning has led to numerous multi-view stereo~(MVS) methods that leverage this technology. These methods have shown significant potential in terms of accuracy and efficiency, particularly in close-range and aerial reconstruction using pinhole cameras. 
However, current methods primarily concentrate on estimating the depth map in accordance with the fronto-parallel planes of a reference view. These methods are not directly applicable to the RPC model \cite{rpc}, as it lacks explicit physical parameters to define the front of a camera.
To mitigate the differences in imaging geometries between push-broom and pinhole cameras, SatMVS \cite{satmvs} introduces a robust differentiable rational polynomial camera warping~(RPC warping) module. This module enables deep MVS satellite image 3D reconstruction without the need for epipolar rectification.

In SatMVS \cite{satmvs}, several methods have been explored to apply differentiable RPC warping to large-scale MVS reconstruction using remote sensing imagery, such as RED-Net~(RPC) \cite{rednet}, CasMVSNet~(RPC) \cite{casmvsnet}, and UCS-Net~(RPC) \cite{ucsnet}. 
However, these methods primarily adapt the general learning-based MVS framework without considering the specific characteristics of the terrains. These geographical features are pivotal in delineating the earth's surface, and the oversight leads to low accuracy in height estimation. 
From the macro perspective, the terrains are typically undulating, while from the micro perspective, the terrains can be defined as composed of countless small planes. In the context of remote sensing height estimation, previous works mainly faced two issues when terrain characteristics were not considered: (1)~using equal interval partition might not effectively cover the undulating surface, resulting in inaccurate height estimations; (2)~estimating the height for each pixel without taking into account the surrounding pixel height values might cause the estimated height are not in an effective height range for a single plane, i.e., there may be an abnormal height value, which leads to inaccurate height estimation. To address these issues, we propose using slope to measure the Earth's surface, with subsequent discussions centered around this approach.
% Specifically, at the macro level, the terrains are usually undulated, so the above methods may not be effective in covering undulating terrains using equal interval partition, resulting in inaccurate height estimation.
% Conversely, at the micro level, the terrains can be defined as composed of countless small planes, but the previous methods estimate the height for each pixel without taking into account the surrounding pixel height values, so there may be estimated error values that are not in an effective height range for a single plane, i.e., there may be an abnormal height value, which leads to inaccurate height estimation.

% 从宏观上来看，地表通常是起伏不定的，反之从微观上来看，地形中可以定义为由无数个微小平面组成。针对遥感高度重建，之前的工作主要存在两点问题：（1）采用等间隔划分可能并不能够有效地涵盖起伏不定的地表，从而造成不准确的高度估计。（2）反之再针对逐像素进行高度估计时，由于微小的平面可能会存在高度不在一个高度范围的情况，即存在异常高度估计，进而造成高度估计不准确。我们为了解决上述问题提出了利用坡度来对地表进行度量，整体后面的内容介绍围绕坡度来讲。
% 从宏观上来看，地表通常是起伏不定的，因此上述方法采用等间隔划分可能并不能够有效地涵盖起伏不定的地表，从而造成不准确的高度估计
% 反之从微观上来看，地形中可以定义为由无数个微小平面组成，之前的方法针对逐像素进行高度估计，因此针对微小的平面可能会存在高度不在一个高度范围的情况，即存在异常高度估计，进而造成高度估计不准确。
% 
% 地表通常是起伏不定的，从这里指出前者方法的在遇到这种情况的时候，可能会出现XXX问题（这里的问题要讲清楚），我们为了解决该问题提出了对于坡度来对地表进行度量，整体后面的内容介绍围绕坡度来讲。

% 坡度是地理学中一个重要的的概念，它可以用来反映地表的稳定性和流动性，其中坡度通常用来衡量地表单元陡缓的程度。因此结合上之前所提到的，我们可以假定在遥感图像中地表是由无数个微小的坡面组成的。

The slope is the fundamental characteristic of the earth's surface to reflect its stability and mobility, and other high-level terrain characteristics are mainly developed based on slope measurement \cite{slope, abramson2001slope}. Therefore, incorporating slope awareness into a MVS pipeline appears essential for achieving effective large-scale reconstruction of the earth’s surface. Slope, in essence, corresponds to the concept of gradient in mathematics, where it quantifies the rate of change of a variable with respect to another \cite{liu1994slope}. In remote sensing imagery, we can analogously view the earth's surface as comprising numerous small slope surfaces, akin to computing gradients in mathematical functions. By examining how terrain variables change over small distances in various directions, we can infer the slope of the surface. Specifically, in high-resolution remote sensing images, we can assume that a 3×3 pixel plane is the smallest slope surface, and the height values of a single slope surface will not change extremely. Thus the slope for a given pixel can be defined by the difference between the height value of that pixel and the maximum of the height values of the eight neighboring pixels around it, while the slope direction is defined as the direction from the maximum height value to the center pixel. As shown in Figure \ref{fig:figure_1}, we can utilize above strategy to obtain the slope direction map based on the height map. This map can then be utilized to refine reconstruction results at both micro and macro levels, offering enhanced accuracy and detail in the reconstructed terrain.

% 具体地，在高分辨率的遥感影像中我们假设3×3像素的平面为最小的坡面，并且单个坡面的高度值不会发生图片，因此对于某个像素的坡度值可以通过该像素的高度值和周围八个邻近像素高度值中的最大值的差值来定义，同时坡度方向定义为从最大高度值到中心像素的方向。

% 将一下坡度是地理信息学中非常常见的用来描述地形变化的概念（详细调研一下），然后受启发于这个特性，我们可以定义如下假设：微观小平面的高度变化趋势是相同的，即单个平面的高度差值并不会发生突变（即平面是真平的，但是不是水平，可以是倾斜的），
% 为了解决上述，我们提出了一种基础高度图计算坡度图以此来衡量地表变化的方法。介绍一下方法的，首先是基于微观概念的假设，然后将高度图同样看做无数个小平面（3×3），并且认为每个平面的高度变化趋势是相同的，这样我们将中间值和该平面的最大高度值之间的差值看作是该平面的坡度，同时将最大值到中间值的方向作为该平面的坡度方向。这里最好给个图例，可视化一下坡度方向图

Motivated by the above thoughts, we proposed TS-SatMVSNet, an end-to-end slope-aware framework for large-scale remote sensing terrain reconstruction. Drawing from mathematical gradient concepts, we first compute slopes within small pixel planes and then utilize slope direction maps to enhance reconstruction outcomes in three key aspects. Firstly, in order to constrain the framework, we proposed a slope direction loss between the predicted slope direction map and the pseudo ground truth (GT) slope direction map for implicitly optimizing height estimation results, where the slope direction map is generated by the corresponding height map. Secondly, at the macro level, benefiting from the degree of undulation of the surface that can be measured by the slope, we propose a slope-guided interval partition module by adaptively adjusting the pixel-wise height interval to refine height estimation using slope values. Finally, at the micro level, we design a height correction module by using a 3×3 learnable Gaussian smoothing operator for each small plane to amend inaccurate height values to achieve more accurate height estimation. 
In summary, by introducing slope and slope direction into the MVS pipeline, our model can effectively fit the complex terrain undulated trend in the remote sensing domain by calculating the slope direction map and being constrained by the slope direction loss. Furthermore, our model has combined two slope-guided modules proposed from the macro and micro levels to further achieve more accurate terrain height estimation.
The ablation study has also validated the effectiveness of our proposed modules.
Moreover, extensive experiments on several benchmark datasets (e.g.WHU-TLC dataset \cite{satmvs}, MVS3D dataset \cite{mvs3d}, US3D dataset \cite{US3D}) demonstrate that our approach achieves excellent performance and demonstrates competitive generalization ability compared to all-listed methods. Specifically, the results as far as correctness and accuracy exceeded the results of other SatMVS-based methods in a between-method comparison by at least 16\% in MAE metric and at least 5\% in $< 2.5m$ metric at WHU-TLC dataset. % 还需要写一下其他数据集上的结果

% 通过将坡度和坡度方向引入到MVS pipeline中，我们的模型针对复杂的地形场景，能够通过高度度计算坡度图并受到坡度约束来有效地拟合遥感影像中的地形变化趋势，进而结合分别从宏观和微观提出的两个模块来进一步实现高精度的地形高度估计。

Our main contributions are summarized as follows:
\begin{itemize}
    \item We propose an end-to-end slope-aware framework TS-SatMVSNet for height for large-scale remote sensing terrain reconstruction.
    \item We propose a height-based slope calculation strategy to measure terrain undulation and provide a visualization method for the slopes.
    \item We propose two slope-guided modules from macro- and micro-levels to incorporate the slop information into the MVS pipeline to achieve more accurate height estimation.
    \item We propose a slope direction loss to constrain the model to improve the accuracy of height estimation in the remote sensing MVS domain.
\end{itemize}

\section{Related Work}
\subsection{Classical Stereo Algorithms for 3D Earth Surface Reconstruction}
3D reconstruction of the Earth’s surface from satellite imagery is primarily accomplished through conventional geometric methods, which can be broadly categorized into two main types. The first type is grounded in the epipolar geometry of satellite images, with the RPC Stereo Processor (RSP) \cite{rsp} serving as a prime example. In this approach, stereo images are initially rectified in accordance with the RPC model. Subsequently, a stereo matching algorithm such as the semi-global matching (SGM) \cite{sgm} method is employed to estimate disparities. CATENA \cite{catalyst} employs SGM in conjunction with distributed optimization to automatically generate a high-resolution digital surface model (DSM). Ultimately, these disparity maps are transformed into 3D points within the world coordinate system. 
The second type involves adapting a complex RPC model into a pin-hole model within a confined area, followed by the application of the stereo/MVS pipeline for reconstruction. Satellite stereo pipeline (S2P) \cite{s2p} rectifies stereo images but approximates the push-broom geometry of small cropped image tiles by a pinhole model, and then performs standard stereo matching. Adapted COLMAP \cite{colmap} uses plane sweeping to avoid epipolar resampling, to reconstruct the 3D structure from multi-view satellite images.
However, these classical methods typically depend on traditional feature extraction operators \cite{sift} to extract features for stereo matching. These operators may not be fully capable of capturing the features of complex terrain surfaces. This limitation could lead to error height estimation, ultimately causing the reconstruction holes.

% 上述立体匹配算法通常是采用传统的特征提取孙子，因此可呢个

\begin{figure*}[!t]
    \centering 
    \includegraphics[width=1\linewidth]{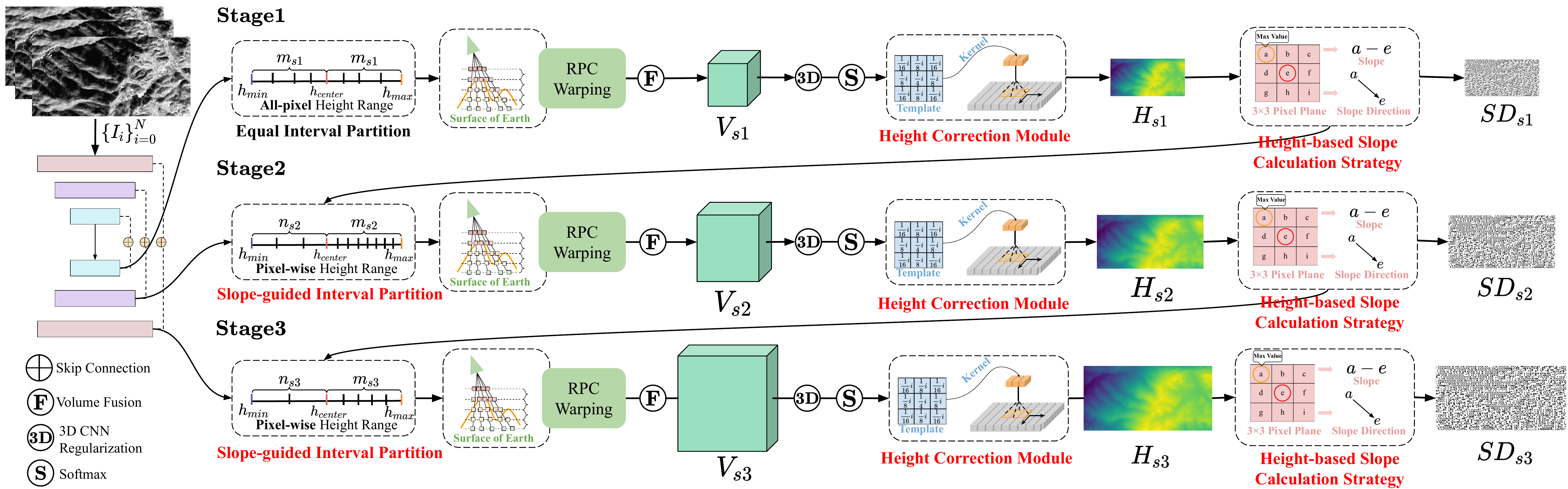}\vspace{-3mm}
    \caption{Illustration of overall TS-SatMVSNet. This is a typical multi-stage coarse-to-fine framework. The modules labeled in orange text are the novel modules we propose in this paper. The concepts of `all-pixel' and `pixel-wise' are given by previous related work \cite{arai-mvsnet}. The details of Height Correction Module and Height-based Slope Calculation Strategy are shown in Fig. \ref{fig:figure_3} and Fig. \ref{fig:figure_5}.}
    \label{fig:figure_2} 
\end{figure*}

\subsection{Deep Learning Based Multi-View Stereo for 3D Earth Surface Reconstruction}
With the development of the artificial intelligence, many deep learning-based Multi-View Stereo~(MVS) methods \cite{mvsnet, rmvsnet, p-mvsnet, fast-mvsnet} have been proposed to overcome the blemish of traditional methods \cite{li2015detail,seitz2006comparison, stereopsis2010accurate,sun2017robust} by utilizing the learnable convolution. 
MVSNet \cite{mvsnet} firstly proposes an end-to-end MVS framework that extracts features from multiple views by CNNs, thereby achieving high-accuracy reconstruction. P-MVSNet \cite{p-mvsnet} proposes a hybrid 3D U-Net to infer a probability volume from the cost volume and estimate the depth maps. R-MVSNet \cite{rmvsnet} uses convolutional GRUs instead of 3D CNNs to regularize the 2D cost maps. Fast-MVSNet \cite{fast-mvsnet} proposes a sparse cost volume and a Gauss-Newton layer to obtain the high-resolution depth map. MVSNet++ \cite{mvsnet++} introduces a depth-based attention mechanism designed to produce smoother depth maps. EI-MVSNet \cite{ei-mvsnet} employs an epipolar-guided volume construction approach for depth map prediction. NR-MVSNet \cite{nr-mvsnet} implements the DHNC aimed at gathering more promising depth hypotheses, which enhances the DRRA modules' ability to predict more accurate depth maps.

However, current methods primarily depend on the pin-hole camera model to estimate the depth map in accordance with the fronto-parallel planes of a reference view. So it is difficult to directly apply the state-of-the-art deep learning-based MVS methods to satellite imagery with a complex RPC model. 
Therefore, SatMVS \cite{satmvs} attempts to fill this gap by proposing a general deep learning-based MVS framework for satellite images. Specifically, SatMVS proposes a differentiable RPC warping module to apply the SOTA learning-based MVS technology to the satellite MVS task for large-scale Earth surface reconstruction, such as RED-Net (RPC) \cite{rednet}, CasMVSNet (RPC) \cite{casmvsnet}, and UCS-Net (RPC) \cite{ucsnet}. Sat-MVSF \cite{satmvsf} extends SatMVS \cite{satmvs} to provide a comprehensive description of each step involved when modern deep learning-based technology is applied to the task of 3D reconstruction of satellite imagery. RS-MVSNet \cite{rs-mvsnet} proposes a deep learning-based framework to infer a digital surface model from multi-view optical remote sensing images.

However, these methods primarily adapt the general learning-based MVS framework without considering the specific characteristics of the terrains. This oversight leads to low accuracy in height estimation.
In this paper, to overcome this challenge, we attempt to incorporate the characteristics of terrains by proposing an end-to-end slope-aware height estimation network for large-scale remote sensing terrain reconstruction.

\section{Methodology}

\subsection{Pipeline}
The overall framework of TS-SatMVSNet is shown in Fig.\ref{fig:figure_2}. It employs a three-stage coarse-to-fine framework for terrain height estimation.
In stage 1, we execute standard MVSNet pipeline to obtain height map, then calculate the slop information based on the height map for later slope-guided interval partition of stage 2/3. And the height correction module adopted across three stages.
Specifically, prior to entering the pipeline, a FPN \cite{fpn} is used to extract multi-scale context terrain surface features $\left\{I_{i}\right\}_{i=0}^{N}$ from $N$ input images.
\begin{itemize}
    \item  \textbf{Stage 1:} we use the RPC warping \cite{satmvs} based on the all-pixel height range to construct the cost volume $V_{s1}$. And after the regularization and softmax, we adopt a height correction module to amend inaccurate height values for predicted height map $H_{s1}$. Then we follow the height-based slop calculation strategy to obtain the slope map $S_{s1}$ and slope direction map $DS_{s1}$.
    \item  \textbf{Stage 2:} we utilize the slope-guided interval partition based on the slop map $S_{s1}$ and pixel-wise height range to achieve 
    adaptively adjusting the pixel-wise height interval. Similar to Stage 1, after obtaining the cost volume $V_{s2}$, we also adopt the height correction module and height-based slop calculation strategy to process the corresponding item.
    \item  \textbf{Stage 3:} Stage 3 is similar to Stage 2, but with a greater number of height hypothesis planes and a larger scale feature map for improved performance.
\end{itemize}
% loss introduction
After obtaining the height maps and slop direction maps, we adopt the smooth L1 loss \cite{fast_r-cnn} and our proposed slope direction loss to constrain the model.

\begin{figure}[!t]
    \centering 
    \includegraphics[width=1\linewidth]{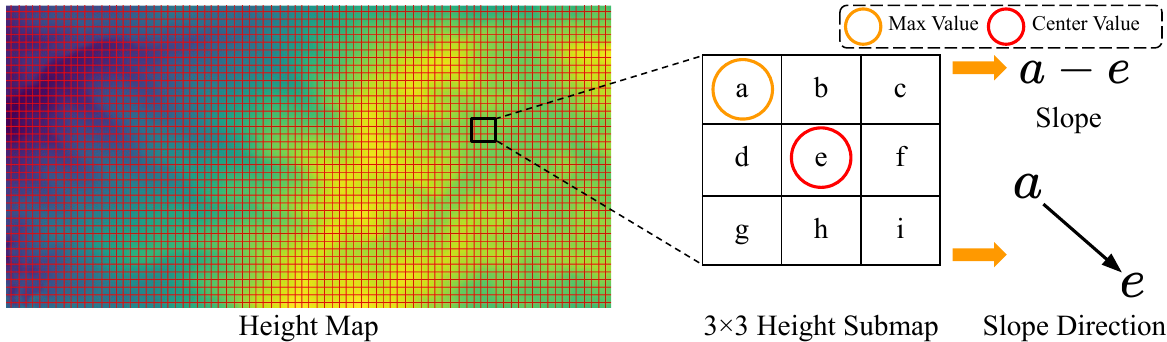}\vspace{-3mm}
    \caption{Illustration of Height based Slope Calculation Strategy.}
    \label{fig:figure_3} 
    \vspace{-3mm}
\end{figure}

\subsection{Height based Slope Calculation Strategy}
The Earth's continental regions are predominantly characterized by undulating terrains, with slope serving as a fundamental parameter to depict their undulation, akin to the mathematical concept of gradient \cite{liu1994slope}. To incorporate the slope into our MVS pipeline, we assume the terrain as comprised of numerous small planes, each exhibiting slope at a micro level, analogous to the concept of gradient computation. This assumption can transform the Earth's surface in remote sensing images into a multitude of small slope planes, enabling slope awareness in our MVS pipeline. Based on the assumption, we consider adopting 3×3 pixel planes as the smallest plane in the deep learning-based MVS framework. Then we propose a height-based slope calculation strategy based on above assumption to obtain the slope map and slope direction map without introducing additional supervised information to improve the performance of height estimation.

Specifically, as shown in Fig \ref{fig:figure_3}, our strategy can convert the height map to slope map and slope direction map. In order to implement the above strategy, we have defined two calculation criteria for slope and slope direction respectively. Firstly, for the slope calculation, we generate corresponding 3×3 pixel smallest planes for each pixel~(achieved by Unfold function of PyTorch \cite{pytorch}). Then, for each smallest plane $p^{3\times 3}(x)$, we take the absolute value of the difference between the maximum height $max(p^{3\times 3}(x))$ and the height value $H(x)$ of the center pixel as the slope value $s=S(x)$ corresponding to that pixel $x$. The formulation of this process as Eq. \ref{eq:eq_1}:

% 为了实现上述策略，我们分别定义了坡度和坡度方向的计算准则。首先对于坡度计算准备，我们针对每个像素都生成对应3×3 pixel smallest planes，然后针对每个最小平面我们其中最大高度和中心像素的高度值的差值的绝对值作为该像素对应的坡度值。

\begin{equation}
 \label{eq:eq_1}
\begin{aligned}
    &p^{3\times 3}(x)=\begin{bmatrix} a & b & c \\ d & e & f\\ g & h & i \end{bmatrix} \\
    &S(x) = | max(p^{3\times 3}(x)) - e | \\
\end{aligned}
 \end{equation}

Secondly, as shown in Figure \ref{fig:figure_3}, the slope direction represents the direction from the center pixel to the maximum height. In order to effectively represent the slope direction, we have established a set of directional codes implemented by PyTorch \cite{pytorch} where adopts the numbers to represents the directions. Specifically, since the size of our smallest plane is 3×3, there are a total of 9 slope directions, which are lower right: 0, down: 1, lower left:2, right:3, vertical:4, left:5, upper right:6, up:7 and upper left:8. Among them, vertical corresponds to the height value of the center pixel being the maximum height, as shown in Fig \ref{fig:figure_4}. The exact procedure regarding the slope direction calculation algorithm we describe using pseudo-code:
% 如图3所示，坡度方向代表的是中心像素到高度最大值的方向。为了有效地表示坡度方向，我们制定了一套方向编码。具体得，由于我们最小平面的大小为3×3，因此总共有9个坡度方向，分别为左上、上、右上、右、右下、下、左下、左以及垂直，其中垂直为高度最大值为中心像素对应的高度值。关于坡度方向的具体过程我们使用伪代码进行描述：
\begin{figure}[!t]
    \centering 
    \includegraphics[width=0.55\linewidth]{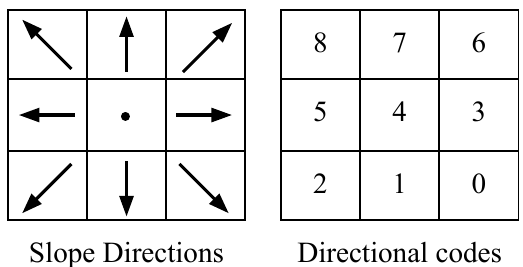}\vspace{-3mm}
    \caption{Illustration of Slope Directional Codes.}
    \label{fig:figure_4} 
\end{figure}

% 这里写一下计算的伪代码:
\begin{algorithm}[H]
    \SetAlgoLined
    \SetKwInOut{Input}{input}
    \SetKwInOut{Output}{output}
    
    \Input{Height Map}
    \Output{Slope Direction Map}
    
    initialization\; $p^{3\times 3}$ $\leftarrow$ Unfold(Height Map)\;
    
    $index_{max}$ $\leftarrow$ Max($p^{3\times 3}$) for each $3 \times 3$ plane \;
    
    $S_{diff}$ $\leftarrow$ $p^{3\times 3}$[$index_{max}$] - Height Map \;
    
    \If{$S_{diff}$ $<$ 0}{
      index1 $\leftarrow$ $S_{diff}$ $<$ 0 \;
      
      $S_{diff}$[index1] $\leftarrow$ -8 \;
      
    }\;
    \If{$S_{diff}$ $>$ 0}{
    index2 $\leftarrow$ $S_{diff}$ $>$ 0 \;

    $S_{diff}$[index2] $\leftarrow$ 0 \;
    
    }\;
    Slope Direction Map $\leftarrow$ Abs($S_{diff}$ $+$ $p^{3\times 3}$[$index_{max}$]) \;
    \caption{Slope Direction Calculation Algorithm}
\end{algorithm}

Through the above two calculation criteria, we can generate the slope map and slope direction map by the height map. The slope map is used in the Slope-guided Interval Partition module. And the slope direction map is used in the Slope Direction Loss, which allows for self-supervised constraint without introducing additional data. In addition, the slope direction map can also be used for visualization, as shown in Fig \ref{fig:figure_1}.
% 通过上述两种计算规则，我们可以通过高度图来计算坡度图和坡度方向图，其中坡度图回用于Slope-guided Interval Partition模块中，而坡度方向图用于Slope Direction Loss中，实现在不引入额外数据的情况下进行自监督约束，此外坡度方向图还可以用于可视化，如图1所示。
% 结尾指出，计算得到的slope map会用于Slope-guided Interval Partition，slope direction map会用于监督

\subsection{Slope-guided Interval Partition Module}
In the remote sensing domain, most existing MVS methods \cite{satmvs, rs-mvsnet} employ equal interval partition for pixel-wise height estimation, which may not effectively cover undulating terrain, ultimately resulting in imprecise height estimation. To address this issue, we propose a slope-guided interval partition module, which utilize the slope of each pixel to reallocate the distribution of height hypothesis planes of each pixel. This module desires to allocate more dense height hypothesis planes within the height range with larger slopes, vice versa, relatively sparse planes within the range with smaller slopes. This is done in order to accurately estimate the height of undulating terrain.

%先写一下我们follow之前的方法，计算逐像素的高度范围，写一下具体如何计算的。然后不同于之前方法采用等间隔划分，我们这里使用的是slope-guided interval partition，在去详细介绍如何划分的。
Before introducing the height hypothesis plane partition strategy, we firstly introduce the calculation process of the pixel-wise height range. Following the previous MVS methods \cite{satmvs, ucsnet, arai-mvsnet}, we also adopt the similar calculated criteria to obtain the pixel-wise height range. Specifically, we utilize previous stage height map $H$ and probability volume $P$ to calculate the pixel-wise standard deviation $\hat{\sigma}(x)$ , which is defined as Eq. \ref{eq:eq_2}:

\begin{equation}
    \label{eq:eq_2}
    \hat{\sigma}(x)=\sqrt{\sum_{m}^{M} P_{m}(x)  \cdot\left(d_{j}(x)-H(x)\right)^{2}}
\end{equation}

\noindent where $M$ is the total number of the height planes, $d_{j}(x)$ represents $j^{th}$ plane of height hypothesis planes of pixel $x$. Then we leverage above results to calculate the pixel-wise height range, the upper and lower boundaries of the pixel-wise height range defined as Eq. \ref{eq:eq_3}:

\begin{equation}
    \centering
    \label{eq:eq_3}
    \begin{aligned}
        &H_{min}(x) = H(x) - \hat{\sigma}(x), \\
        &H_{max}(x) = H(x) + \hat{\sigma}(x) \\
    \end{aligned}
\end{equation}

Based on the above upper and lower boundaries of the pixel-wise height range, we propose a slope-guided interval partition to reallocate the height hypothesis planes for each pixel. 
Specifically, given a height map $H$, we firstly utilize the slope calculation criteria to respectively obtain the slope factors of each pixel, i.e calculate the different $S_{max}(x)$ between the maximum height value and the center pixel $H(x)$ in the 3×3 plane $p^{3\times 3}(x))$, as well as the difference $S_{min}(x)$ between the minimum height value and the center pixel $H(x)$. Then we assign $S_{max}(x)$ as the weight for the range from $H(x)$ to $H_{max}(x)$, and $S_{min}(x)$ as the weight for the range from $H(x)$ to $H_{min}(x)$. Next, we use Eq. \ref{eq:eq_4} to distribute $M$ to the two pixel-wise subranges.

\begin{equation}
    \centering
    \label{eq:eq_4}
    \begin{aligned}
        & M_{l2c}(x) = M \times \frac{S_{min}(x)}{S_{min}(x) + S_{max}(x)} \\
        & M_{c2u}(x) = M \times \frac{S_{max}(x)}{S_{min}(x) + S_{max}(x)} \\
    \end{aligned}
\end{equation}

Thus, we leverage the $M_{l2c}(x)$ and $M_{c2u}(x)$ to respectively calculate the corresponding height interval of the pixel-wise subrange following the Eq. \ref{eq:eq_5}:

\begin{equation}
    \centering
    \label{eq:eq_5}
    \begin{aligned}
        & I_{l2c}(x) = \frac{H_{min}(x) - H(x)}{M_{l2c}(x)} \\
        & I_{c2u}(x) = \frac{H_{max}(x) - H(x)}{M_{c2u}(x)} \\
    \end{aligned}
\end{equation}

Finally, we can obtain the reallocated height hypothesis planes to achieve the slope-guided interval partition, as defined by Eq. \ref{eq:eq_6}.

\begin{equation}
    \centering
    \label{eq:eq_6}
    \begin{aligned}
        & [H_{min}(x), ..., H_{min}(x)+I_{l2c}(x)\times i, ..., H(x)] \\
        & [H(x), ..., H(x)+I_{l2c}(x)\times j, ..., H_{max}(x)] \\
    \end{aligned}
\end{equation}

\noindent where $i$ is the enumerated value of $M_{l2c}(x)$, $j$ is the enumerated value of $M_{c2u}(x)$.
% 这个模块希望能够在坡度更大的高度区间内分配更多地高度假设平面，同时相反地在坡度小的区间内分配更少的平面。
% 大多数方法采用通用的等间隔划分，但是划分的间隔可能并不能够有效地覆盖起伏不平的地形。

\subsection{Height Correction Module}
\begin{figure}[!t]
    \centering 
    \includegraphics[width=0.95\linewidth]{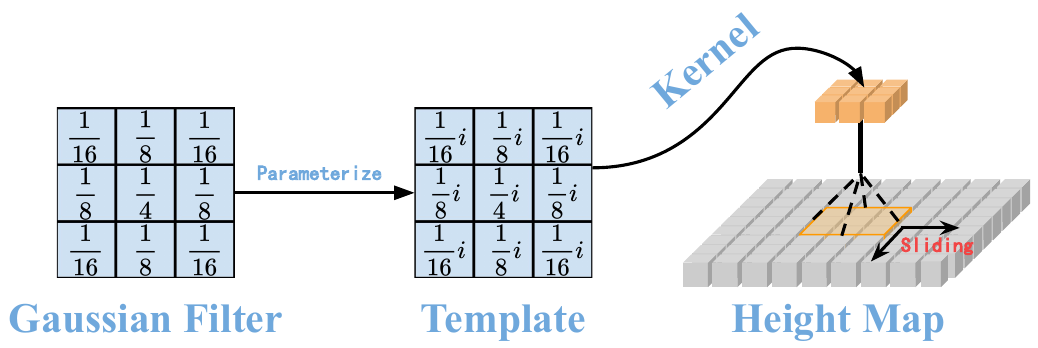}\vspace{-3mm}
    \caption{Illustration of Height Correction Module.}
    \label{fig:figure_5} 
\end{figure}

% 正如我们知道的，高斯滤波是一种线性平滑滤波，适用于消除高斯噪声，广泛应用于图像处理的减噪过程 \cite{}。
The Earth's surface in remote sensing images is composed of countless small slope faces at micro level. Thus the height values of a single slope surface will not change extremely. As we know, Gaussian Filter is a linear smoothing filter, suitable for eliminating Gaussian noise, and is widely used in the noise reduction process of image processing  \cite{gaussian}. Since the Gaussian Filter is unlearnable, and its fitting ability is not powerful enough for height estimation of remote sensing images with a large number of data. Furthermore, considering the similarity between the structure of the Gaussian Filter and a convolutional kernel. Inspired by above mechanism, we propose a height correction module by leveraging a 3×3 learnable Gaussian Filter for each small slope surface to amend inaccurate height value.
Specifically, as shown in Fig. \ref{fig:figure_5}, we firstly construct a general Gaussian Filter with considering the characteristic of height estimation. Then we parameterize the Gaussian Filter as the kernel $K$ of the convolution module to act on the height map to amend the abnormal height noise. The formulation of our height correction module is defined as Eq. \ref{eq:eq_7}:

\begin{equation}
    \centering
    \label{eq:eq_7}
        \begin{aligned}
            & K = \left\{\frac{1}{16}i, \frac{1}{8}i, \frac{1}{16}i, \frac{1}{8}i, \frac{1}{4}i, \frac{1}{8}i, \frac{1}{16}i, \frac{1}{8}i, \frac{1}{16}i \right\} \\
            & \tilde{H} = K \odot H \\
        \end{aligned}
\end{equation}

\noindent where $\tilde{H}$ is the result of being amended.

\subsection{Loss Function}
To obtain the high-quality outputs of each stage, we adopt two loss functions: Height Map Loss (for height map), Slope Direction Loss (for slope direction map).

\subsubsection{Height Map Loss}
Height Map Loss is adopted to measure the difference between the GT height map and the predicted height map to constrain the height estimation. Following the previous methods \cite{satmvs, arai-mvsnet}, we adopt same mean absolute difference loss as our loss:

\begin{equation}
    \label{eq:equation_8}
    \begin{aligned}
    L_{h} =\sum_{i=1}^{3} \lambda_{i} \sum_{x \in \mathbf{x}_{\text {valid }}} \|\hat{H}_{s_{i}}(x)-H_{s_{i}}(x)\|_{1}, \\
    \end{aligned}
\end{equation}

\noindent where $ \hat{H}^{s_{i}} $ denotes the GT height map of each stage, $ H^{s_{i}} $ denotes the predicted height map of each stage. We set $\lambda_{i}$ to be 0.5, 1.0, 2.0 for each stage, $p_{\text {valid }}$ denotes the valid point set of the GT height map.

\subsubsection{Slope Direction Loss}
Slope Direction Loss is designed to measure the difference in the slope direction map between the prediction and the pseudo ground truth to constrain our model effectively capture the undulating trend of terrain. However, since current mainstream remote sensing MVS datasets lacks the ground truth~(GT) slope direction map, it is difficult to directly calculate the loss term by the slope direction map. Thus, we consider generating a pseudo GT slope direction map $SD$ from the GT height map $\hat{H}$ to participate in the loss calculation. Specifically, we adopt our proposed height based slope calculation strategy to generate the pseudo GT slope direction map. 
The Slope Direction Loss is defined as the mean squared error (MSE: L2 distance) between the predicted edge map and the pseudo GT slope direction map. By using this loss function, we can ensure that our model is effectively capture the undulating trend of terrain and producing results that are consistent with nearly real-world slope direction maps. The formulation of Slope Direction Loss is defined as follows:

\begin{equation}
 \label{eq:equation_9}
\begin{aligned}
    L_{s} =\sum_{i=1}^{3} \lambda_{i} \sum_{x \in \mathbf{d}_{\text {valid }}} \|\hat{SD}_{s_{i}}(x)-SD_{s_{i}}(x)\|_{2}, \\
\end{aligned}
 \end{equation}

\noindent where $ \hat{SD}^{s_{i}} $ denotes the GT slope direction map of each stage, $ SD^{s_{i}} $ denotes the generated slope direction map of each stage. We set $\lambda_{i}$ to be 0.5, 1.0, 2.0 for each stage, $d_{\text {valid }}$ denotes the valid point set of the GT slope direction map.

\subsubsection{Overall Loss}
By using the weighted sum of the aforementioned loss terms, we create a comprehensive training criterion for our network. This approach enables us to optimize the network parameters through backpropagation. As a result, our network can learn to produce accurate and robust height maps by minimizing the overall loss, which is a crucial factor in achieving high performance in height estimation tasks.

 \begin{equation}
 \label{eq:equation_10}
\begin{aligned}
    L_{overall} = \lambda_{1} L_{h} + \lambda_{2} L_{s} \\
\end{aligned}
 \end{equation}

\noindent where $\lambda_{1}=0.5, \lambda_{2}=0.5$ are hyper-parameters empirically set based on our experiments on the validation set.

\section{Experiment}
\subsection{Satellite MVS Datasets}
In the realm of satellite MVS using deep learning, we are presently confronted with a substantial scarcity of training datasets. To our understanding, the only relevant datasets available for this purpose are WHU-TLC \cite{satmvs}, MVS3D \cite{mvs3d}, and US3D \cite{US3D}. However, the US3D dataset \cite{US3D} is aimed at the joint task of semantic segmentation and 3D reconstruction, where the scene variations between the stereo image pairs are not suitable for high accuracy MVS reconstruction, so we opted to use WHU-TLC \cite{satmvs} and MVS3D \cite{mvs3d} dataset in our pipeline.

\noindent \textbf{WHU-TLC dataset}: WHU-TLC \cite{satmvs} is a comprehensive satellite multi-view dataset, comprising triple-view images captured by the TLC camera on the Ziyuan-3 (ZY-3) satellite. This dataset includes a multitude of image patches, each accompanied by their respective RPC parameters and corresponding height maps. These height maps are derived from projecting the DSMs onto the images using the RPC parameters. Although theoretically similar to the depth map in a close-range MVS dataset, the height map stores the height information of the corresponding pixel in the image, rather than the depth. Each 5120×5120 image is segmented into 768×384 patches, with a 5\% overlap in both horizontal and vertical directions. The dataset contains a total of 5011 training sets.

\noindent \textbf{MVS3D dataset}: The MVS3D dataset is a multi-view stereo benchmark for satellite imagery. It comprises 50 WorldView-3 images and airborne LiDAR data, which are used to establish the ground truth. However, the RPC parameters have not been calibrated, leading to a lack of geometric consistency between the matched point clouds and the ground truth. The panchromatic image’s GSD is approximately 0.3 m. The images were acquired over a period spanning from November 2014 to January 2016, while the ground truth data was collected in June 2016. Consequently, significant scene differences exist both among the stereo images and between the images and the ground truth. This implies that reconstructing these scenes poses a formidable challenge, particularly for methods based on deep learning. Furthermore, the MVS3D dataset is considerably smaller than the WHU-TLC dataset and does not provide sufficient training samples for deep learning-based methods.

\subsection{Evaluation Metrics}
\label{evaluation_metrics}
In this paper, we adopt the following metrics~(similarly adopted in SatMVS\cite{satmvs}) to evaluate the quality of DSMs in different datasets:

\noindent \textbf{WHU-TLC Dataset}
\begin{itemize}
    \item MAE: the average of the $L_{1}$ distance over all the grid units between the ground truth and the estimated DSM, the formulation is defined as Eq. \ref{eq:equation_11}:
        \begin{equation}
        \label{eq:equation_11}
        \begin{aligned}
            M A E=\frac{\sum_{(i, j) \in D \cap \widetilde{D}}\left|h_{i j}-\tilde{h}_{i j}\right|}{\sum_{(i, j) \in D \cap \tilde{D}} Iver(i, j)}
        \end{aligned}
        \end{equation}
    \noindent where $D$ and $\widetilde{D}$ represent the valid grid cells in the estimated DSM and ground truth, $h_{i j}$ and $\tilde{h}_{i j}$ refer to the height value of the estimation and ground truth in the grid cell in row $i$ and column $j$, and $Iver$ represents the Iverson bracket, which means 1 if $A$ is true and 0 otherwise.
    \item RMSE: the standard deviation of the residuals between the ground truth DSMs and the estimated DSMs, as defined by Eq. \ref{eq:equation_12}:
        \begin{equation}
        \label{eq:equation_12}
        \begin{aligned}
            R M S E= \sqrt{\frac{\sum_{(i, j) \in D \cap \widetilde{D}}\left(h_{i j}-\tilde{h}_{i j}\right)^{2}}{\sum_{(i, j) \in D \cap \tilde{D}} Iver(i, j)}}
        \end{aligned}
        \end{equation}
    \item $<$2.5m: the percentage of grid units with an L1 distance error below the thresholds of 2.5 m, the definition of this formula is shown in Eq. \ref{eq:equation_13}:
        \begin{equation}
        \label{eq:equation_13}
        \begin{aligned}
            < 2.5m=\frac{\sum_{(i, j) \in D \cap \widetilde{D}}Iver(\left|h_{i j}-\tilde{h}_{i j}\right|<2.5)}{\sum_{(i, j) \in D \cap \tilde{D}} Iver(i, j)}
        \end{aligned}
        \end{equation}
    \item $<$7.5m: the percentage of grid units with an L1 distance error below the thresholds of 7.5 m, the definition of this formula is shown in Eq. \ref{eq:equation_14}:
        \begin{equation}
        \label{eq:equation_14}
        \begin{aligned}
            < 7.5m=\frac{\sum_{(i, j) \in D \cap \widetilde{D}}Iver(\left|h_{i j}-\tilde{h}_{i j}\right|<7.5)}{\sum_{(i, j) \in D \cap \tilde{D}} Iver(i, j)}
        \end{aligned}
        \end{equation}
    \item Comp: the percentage of grid units with valid height values in the final DSM.
\end{itemize}

\noindent \textbf{MVS3D Dataset}
\begin{itemize}
    \item RMSE: the standard deviation of the residuals between the ground truth DSMs and the estimated DSMs, as defined by Eq. \ref{eq:equation_12}
    \item $<$1.0m: the percentage of grid units with an L1 distance error below the thresholds of 1.0 m, the definition of this formula is shown in Eq. \ref{eq:equation_15}:
        \begin{equation}
        \label{eq:equation_15}
        \begin{aligned}
            < 1.0m=\frac{\sum_{(i, j) \in D \cap \widetilde{D}}Iver(\left|h_{i j}-\tilde{h}_{i j}\right|<1.0)}{\sum_{(i, j) \in D \cap \tilde{D}} Iver(i, j)}
        \end{aligned}
        \end{equation}
    \item Median: the median value of the absolute error between the estimated DSMs and ground truth DSMs in the valid grid cells, as defined by Eq. \ref{eq:equation_16}:
        \begin{equation}
        \label{eq:equation_16}
        \begin{aligned}
            \text { Median }=\underset{(i, j) \in D \cap \widetilde{D}}{\operatorname{median}}\left(\left|h_{i j}-\widetilde{h}_{i j}\right|\right)
        \end{aligned}
        \end{equation}
\end{itemize}

\subsection{Experimental Settings}
In this paper, we primarily adopt two datasets to train and test our model. Specifically, following Sat-MVSF \cite{satmvsf}, we also train our TS-SatMVSNet on the WHU-TLC training dataset \cite{satmvs} and evaluate our pretrained model separately on the WHU-TLC test dataset and the MVS3D dataset \cite{mvs3d}.

\noindent \textbf{WHU-TLC Dataset}: During the training phase, TS-SatMVSNet, a PyTorch-based implementation, is trained on the WHU-TLC-V2 dataset using 2x NVIDIA RTX 3090 GPUs, each with 24 GB of memory. The hyperparameters are configured as follows: the batch size is set to 4, RMSProp is chosen as the optimizer, the network undergoes training for 30 epochs, starting with a learning rate of 0.001, this learning rate is halved after the 10th epoch. And a three-stage hierarchical matching approach is employed to deduce height maps from coarse to fine. For the TLC images, the number of input images, denoted as n, is fixed at 3. The numbers of hypothetical height planes and their corresponding intervals are set to [64, 32, 8] and [$\frac{HR_{h} - HR_{l}}{64}$, 5 m, 2.5 m], respectively, where $HR_{h}$ and $HR_{l}$ separately represent the high bound and the low bound of the height range of the current TLC image. In the testing phase, we utilize Sat-MVSF \cite{satmvsf} designed pipeline to evaluate our pretrained model. We adopt parameters similar to those used during the training phase to infer the height maps. The point clouds generated by these height maps are then incorporated to obtain the Digital Surface Models (DSMs). These DSMs serve as a measure to evaluate the performance of our model by using the evaluation metrics in Sec \ref{evaluation_metrics}.

\noindent \textbf{MVS3D Dataset}: Due to there are not enough training samples, we use the our model pre-trained on the WHU-TLC Dataset to validate the generalization ability of our TS-SatMVSNet on satellite images. 
However, since there is currently no very complete and practical material for the MVS3D Dataset \cite{mvs3d}, under Jian’s \cite{satmvsf} guidance, we started from scratch to process the MVS3D dataset to obtain the MVS3D dataset that can be used for MVS evaluation. The specific process includes: image cropping, rpc camera model generation, bundle adjustment, view selection. Regarding the specific process, we will open source it along with our code.
For specifical experimental setting, we also adopt similar testing configuration to WHU-TLC Dataset, the hyperparameters are configured as follows: the image size is cropped to 3072 × 3072 pixels, the batch size is set to 1, the number of input images, denoted as n, is fixed at 3, the numbers of hypothetical height planes and their corresponding intervals are set to [64, 32, 8] and [$\frac{HR_{h} - HR_{l}}{64}$, 5 m, 2.5 m], respectively.

\begin{figure*}[!t]
    \centering 
    \includegraphics[width=1\linewidth]{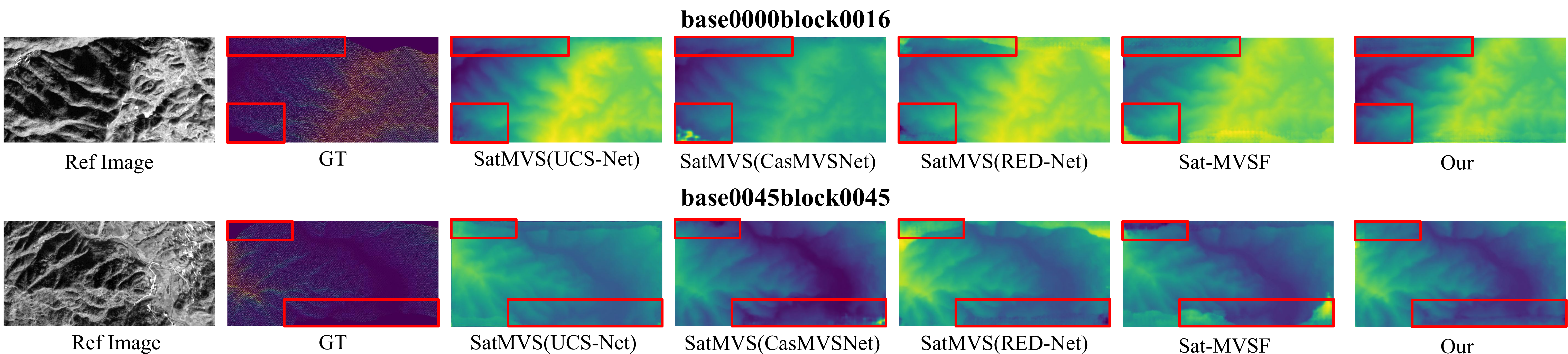}\vspace{-3mm}
    \caption{Qualitative results of mainstream deep learning-based methods on two different areas of the WHU-TLC Dataset. The red boxes in the figure are visualized views of the local details respectively.}
    \label{fig:figure_6} 
\end{figure*}

\input{tlc_mvs_dataset}

\input{tlc_mvs_dataset_1}

\subsection{Quantitative Results on WHU-TLC Dataset}
To demonstrate the effectiveness of our model, we compare our TS-SatMVSNet with two groups of state-of-the-art~(SOTA) methods: traditional MVS methods, e.g., adapted COLMAP \cite{adapted_colmap}, S2P \cite{s2p}, SDRDIS \cite{sdrdis}, ArcGIS \cite{arcgis}, CATALYST \cite{catalyst}, Metashape \cite{metashape}, and deep learning-based MVS methods, e.g., RED-Net~(RPC) \cite{rednet}, CasMVSNet~(RPC) \cite{casmvsnet}, UCS-Net~(RPC) \cite{ucsnet}, SatMVSF \cite{satmvsf}. \par
\noindent \textbf{Comparisons with the traditional MVS methods}: For traditional MVS methods, the quantitative results are shown in Table \ref{tab:Table.2}, we can observe that our method establishes state-of-the-art performance in all metrics by comparing it to the all of the traditional methods. Specifically, our proposed method has improved by 13.3\%, 17.4\%, and 32.6\% on the MAE, RMSE, and $<2.5m$ metrics respectively compared to the best-performing Adapted COLMAP. And we also improved 13.1\% on the $<7.5m$ metric compared to the best-performing CATALYST. We can attribute the superior performance to two special factors:(a)~neural operators demonstrate superior adaptability in multi-scale complex scenes, compared to traditional operators such as SIFT~\cite{sift}; (b)~the incorporation of the terrain slope significantly enhances the scalability and adaptability of our TS-SatMVSNet compared to other traditional pipelines. \par
\noindent \textbf{Comparisons with the deep-learning-based MVS methods}: The quantitative results
for the mainstream deep-learning-based MVS methods are shown in Table \ref{tab:Table.1}. It demonstrates that our method achieves the highest level in various metrics compared to other SOTA methods. For instance, compared to the SatMVS-F \cite{satmvsf}, our approach can improve the MAE from 1.895 to 1.879~(0.8\% performance improvement), the $<2.5m$ from 64.82 to 77.92~(20.2\% performance improvement), and the $<7.5m$ from 80.05 to 97.34~(21.6\% performance improvement) within the $2048 \times 1472$ resolution. Additionally, compared with SatMVS~(RED-Net), although our TS-SatMVSNet exhibits relatively lower performance~(3.654 vs. 3.892), our method still superior SatMVS~(RED-Net) in MAE, $<7.5m$ and Comp respectively, which can improve the MAE from 1.945 to 1.879, $<7.5m$ from 96.59 to 97.34, and Comp from 82.29 to 82.60. We attribute this superior performance improvement to our proposed slope-guided interval partition module, which utilize the slope to reallocate the distribution of pixel-wise height hypothesis planes to obtained more accurate height estimation.
Moreover, we also conducted a comparative analysis of the impact of different resolutions on model performance~(resolution from $2048\times 1472$ to $5120 \times 5120$). From the blue values in the Table \ref{tab:Table.1}, we can observe that our proposed TS-SatMVSNet has the least fluctuation across all metrics, e.g., 0.1\% $\sim$ 0.6\%. On the other hand, RED-Net \cite{rednet}, which also adopts RPC Warping \cite{satmvs}, exhibits significant result fluctuations across all metrics, 0.3\% $\sim$ 15.9\%. It is worth mentioning that the incorporation of the terrain slope significantly enhances the scalability and adaptability of our ST-SatMVSNet, which causes our lowest fluctuations across all metrics.
% 我们也在5120x5120分辨率上进行了不同分辨率对于模型性能影响的对比，我们从Table中蓝色的数值也可以看出，我们提出的TS-SatMVSNet在各个指标上的浮动最小的，而RED-Net由于并没有使用RPC Warping \cite{satmvs}，因此在各个指标上出现了强烈的结果波动。 我们认为这是由于我们能够对于多尺度图像

% 考虑到论文内容长度，下面定性结果可以单独开一节
% 提供一些可视化的对比结果（DSM以及深度图的对比都要有），同时展示一些Slope Map的可视化结果？Slope MAP可视化结果看看放到哪里合适（要不要单开一节和深度图一起来分析？从而来证明添加这个约束的好处）

\begin{figure*}[!t]
    \centering 
    \includegraphics[width=1\linewidth]{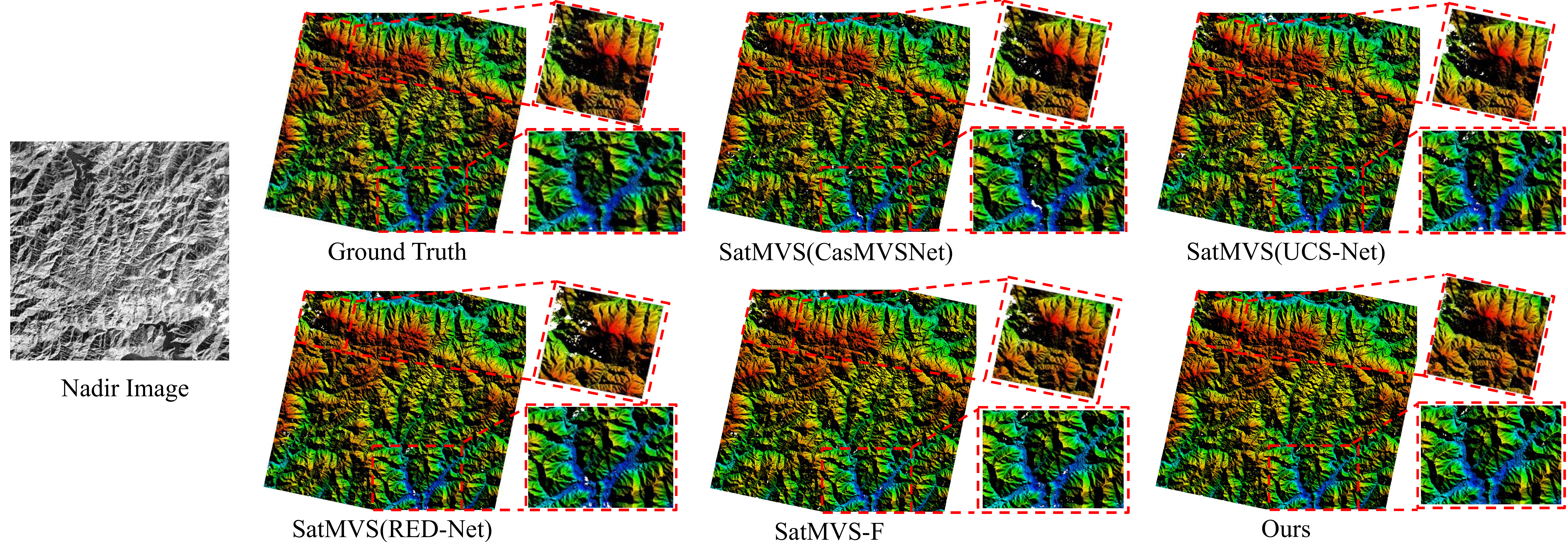}\vspace{-3mm}
    \caption{Visualization examples of the DSM results produced by the different methods on the WHU-TLC dataset. The red dash boxes in the figure are visualized views of the local details respectively.}
    \label{fig:figure_7} 
\end{figure*}

\begin{figure*}[!t]
    \centering 
    \includegraphics[width=1\linewidth]{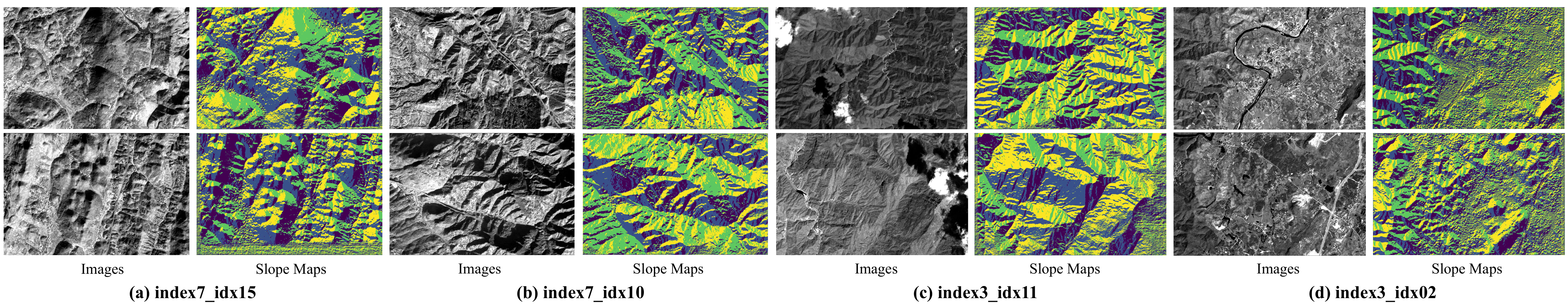}\vspace{-3mm}
    \caption{Visualization examples of the slope maps, which calculated from the predicted height maps by our proposed Height based Slope Calculation Strategy~(HSCS).}
    \label{fig:figure_8} 
\end{figure*}

\subsection{Qualitative Results on WHU-TLC Dataset}
To further verify the effectiveness of our proposed method, we have provided several typical qualitative results on the WHU-TLC dataset. Specifically, we have separately compared the height map and DSM results with the state-of-the-art methods. Firstly, as shown in Figure \ref{fig:figure_6}, we can observe that our method, as compared to current mainstream deep learning methods, exhibits more accurate results in the height estimation of local details. As demonstrated by the red boxes in the figure, other methods experience artifacts in local areas. This benefits from the incorporation of slope-guided information, our method can effectively capture the undulations of the terrain depicted in the image, thereby precisely estimating height values and consequently reducing the occurrence of artifacts.
%我们提出的方法在局部细节的高度估计上相比较于目前主流深度学习的方法展现出了更加准确的结果，如图中红色方框所展示的，其他的方法在局部区域出现了伪影的情况，但是得益于slope-guided information incorporation，我们的方法能够准确的把握图中地势的起伏变化，从而准确地估计出高度值进而减少伪影的产生。
% 下面继续分析DSM的结果
Secondly, we have further illustrated more DSM results to prove the superiority and effectiveness of our proposed method compared to mainstream methods. The qualitative results are shown in Figure \ref{fig:figure_7}, from the comparison of local details in the dashed boxes, it can be seen that our method is capable of reconstructing more complete DSM results compared to the mainstream deep learning methods, e.g. SatMVS, SatMVS-F, which also proves that the more accurate of height estimation of our model can predict. We can attribute this significant improvement is a benefit from our proposed slope-guided interval partition model, which can capture the undulations of the terrain to enhance the ability to perceive terrain changes in order to divide more accurate height intervals and obtain accurate height estimations.

\subsection{Visualization of Slope Map Results}
In the visualization phrase, due to the visual effect of the 10x downsampled slope direction map visualized in the form of a flow map is not intuitive enough (as shown in Figure \ref{fig:figure_1}), it fails to effectively prove the reliability of our proposed Height based Slope Calculation Strategy~(HSCS). Therefore, we have adopted a new visualization scheme that displays by taking the slope values as inputs. The visualization results are shown in Figure \ref{fig:figure_8}, where the slope maps are calculated from the height maps predicted by using our HSCS. 
Benefit from the superior intuitive visualization results, we can obviously observe that our proposed HSCS can effectively capture the undulations of the terrain, thereby integrating the slope information into the pipeline to enhance the perception of the terrain. It is noteworthy that the comparison between slope maps and the original images very visually demonstrates the effectiveness of our proposed HSCS.
Additionally, as shown in Figure \ref{fig:figure_8}~(d), some artifacts are present in the obtained slope map, which can be seen from the correspondence with the original image to be generated from the height estimation results of urban areas. Since our initial design was intended for terrain, these urban areas were not considered, suggesting that future integration of further improvement strategies may be considered.
% 在实际可视化的过程中，由于10x降采样的slope direction map采用flow map的形式进行可视化的视觉效果不够直观~(如Figure 1所示)，导致其不能有效地证明出我们提出的Height based Slope Calculation Strategy~(HSCS)的可靠性，因此这里我们采用了一种全新的可视化代码通过将slop值作为输入来进行展示。 可视化的效果如Figure 8所示，其中图中的slope maps是利用HSCS从我们模型预测的高度图所计算得到的，我们可以观察到我们提出的HSCS可以有效的捕获地形的起伏变换，进而能够将这种坡度信息融入到模型中以此来提升模型对于地形的感知能力。此外值得注意的是，在Figure 8 (d)中，我们得到的slope map存在了一些artifacts的情况，从原图的对应关系可以看出这是城市区域的高度估计结果所生成的。由于我们最初设计的初衷是针对地形来进行设计的，因此对于这些城市区域缺乏考虑，未来可能还需要融合进一步的改进策略。

\subsection{Ablation Study}
% slope-guided的方案可能会导致在城市区域出现效果特别差的情况，因此可能在2.5m的指标上有一定的性能弱势，这里考虑可以单独把所有非城市区域的地形图像来再次进行验证
In this section, we have conducted an ablation study to understand and analyze the contributions of the modules of our architecture. The quantitative results are shown in Table \ref{tab:Table.3} and the qualitative results are shown in Figure \ref{fig:figure_8}. Specifically, our ablation studies are mainly divided into two parts: 
Firstly, we validated the accuracy and advantages of our proposed framework in geographical height estimation. Secondly, we further conducted more ablation analysis to verify the effectiveness of our proposed slope-based modules.
Additionally, we set up the SatMVS~(UCS-Net) as our baseline. 

% \noindent \textbf{Effectiveness of the Height based Slope Calculation Strategy}: 

\noindent \textbf{Effectiveness of the Terrain Height Estimation}: As we know, the WHU-TLC Dataset is a multi-view stereo height estimation dataset that contains many different remote sensing scenes, e.g., terrain, urban, and other areas. However, since our scheme is designed for terrain height estimation, scenes from non-terrain areas may not be suitable for our proposed slope-guided manner, thereby affecting the performance of height estimation. This situation is also reflected in the metric of $<$2.5m in Table \ref{tab:Table.1}, i.e., our method did not obtain the SOTA on this metric. Thus, we can attribute it to the WHU-TLC test set, which includes some urban scene areas and may affects the performance of our method. Therefore, to further verify the effectiveness of our method in the task of terrain height estimation, we constructed a sub-dataset $\text{WHU-TLC}^{*}$ that only includes earth terrain areas based on the existing WHU-TLC test set and conducted comparative experiments, e.g, index7 folders, index3 folders.
The quantitative results are shown in Table \ref{tab:Table.3}, we can observe that using purely terrain images to construct the $\text{WHU-TLC}^{*}$ sub-dataset has resulted in a significant improvement in performance across all metrics. Specifically, MAE improved by 4.6\%, RMSE by 4.1\%, $<$2.5m by 1.4\%, and $<$7.5m by 0.5\%. We attribute it to our novel slope-guided manner, which can effectively capture the undulations of the terrain and integrate slope information into the pipeline, thereby enhancing the performance of terrain height estimation.
% 我们可以观察到在使用了纯地形图像构建$\text{WHU-TLC}^{*}$模型在各个指标上的性能都得到了显著地提升。具体地，MAE提升了4.6\%， RMSE提升了4.1\%, $<$2.5m提升了1.4\%, $<$7.5m提升了0.5\%. 我们将其归因为我们所提出的slope-guided manner，它能够有效地捕获地形的变换并且将坡度信息融合到模型当中，从而提升地形高度估计的性能。
% 正如我们所知道的，WHU-TLC Dataset是一个包含很多不同遥感场景的多视图立体高度估计数据集，因此其主要包含了地形、城市以及其他区域。然而由于我们的方案是针对地形高度估计来设计的，因此非地形区域的数据可能并不适用于我们所提出的slope-guided manner，从而影响高度估计的结果。这种情况也在Table 1中 <2.5m的指标上有所体现, i.e., 我们的方法在该指标上并没有达到SOTA. 我们将其归因于WHU-TLC测试集中包含了一些城市场景区域的重建，可能对于我们的方法的性能有所影响。因此为了进一步验证我们提出的方法在地形高度估计任务上的有效性，我们基于现有的WHU-TLC测试集重建构建了一个仅包含Earth Terrain区域的子集 $\text{WHU-TLC}^{*}$并在这个子集上开展了对比实验。实验结果如Table 3所示.
% 这里先说一下 Table 1中存在<2.5m不是最佳的情况，分析一下是由于上面artifacts的因素导致的，因此我们这里基于现有的WHU-TLC测试集重建构建了一个仅包含Earth Terrain区域的子集 $\text{WHU-TLC}^{*}$, 并在这个子集上开展了对比实验来证明我们的方法能够有效的在地形高度估计任务上展现出十分强大的性能。外加一顿分析xxxx，还可以引入一些Table1中的数据

\input{ablation_study}

\noindent \textbf{Effectiveness of the Slope-guided Interval Partition Module}: Moreover, to verify the effectiveness of our proposed slope-guided interval partition module~(SIPM), we have validated the `baseline + SIPM' on the WHU-TLC dataset, and the resulting metrics are presented in Table \ref{tab:Table.3}. A comparison between Row 1 and Row 2 reveals that the incorporation of slope information to guide the height planes partition, which significantly improves the model’s metrics, including MAE: 2.026 to 1.911, RMSE: 3.921 to 3.898, $<$2.5m: 77.01\% to 77.83\%, $<$ 7.5m: 96.54 \% to 97.15\%. The quantitative results effectively demonstrate the validity of our SIPM.

\noindent \textbf{Effectiveness of the Height Correction Module}: Furthermore, we have added the Height Correction Module~(HCM) with the baseline that involved filtering the slope surface by leveraging 3$\times$3 learnable Gaussian Filter. 
The experimental results are shown in Table \ref{tab:Table.3} Row 1 and Row 3. We observed that upon integrating our HCM, there was a marked improvement in the performance metrics: MAE decreased from 2.026 to 2.002, RMSE reduced from 3.921 to 3.914, $<$2.5m increased from 77.01\% to 77.25\%, and $<$7.5m improved from 96.54\% to 96.90\%. These quantitative results effectively underscore the effectiveness of our HCM.

\subsection{Generalization on MVS3D Dataset}
The MVS3D dataset \cite{mvs3d} serves as the benchmark for the IARPA Multi-View Stereo 3D Mapping Challenge and has been extensively utilized as a standard benchmark in prior methodologies \cite{s2p, catalyst, satmvsf}. To assess the generalizability of our TS-SatMVSNet, we conducted tests using the MVS3D dataset and evaluated the outcomes using the official evaluation scripts. 
% The quantitative results are shown in Table \ref{tab:Table.4}, and we can observe that our method demonstrates highly competitive performance across all sites and achieved state-of-the-art~(SOTA) performance on average evaluation metrics, such as $<$1.0m: 60.635\%, Median: 0.353m, RMSE: 2.898m. 
The quantitative results shown in Table \ref{tab:Table.4} demonstrate that our method demonstrates highly competitive performance across all sites and achieved state-of-the-art~(SOTA) performance on average evaluation metrics, e.g., 60.635\% in $<$1.0m, 0.353m in Median, 2.898m in RMSE.
% For instance, we achieve 60.635\% percent of grid units with an L1 distance error $<$1.0m, and 0.353m for median value of the absolute error between the estimated DSM and ground truth DSM in the valid grid cells, and 2.898m for the RMSE.}
Specifically, we have conducted comparisons with two different types of methods: traditional methods, such as S2P \cite{s2p}, Metashape \cite{metashape}, Adapted COLMAP \cite{adapted_colmap}, and deep learning-based methods, e.g, Sat-MVSF \cite{satmvsf}. Compare to traditional methods, our TS-Sat-MVSNet exhibits SOTA performance on the vast majority of sites. For instance, our method can improve the $<$1.0m from 70.4\%~(S2P) to 71.02\% in Site2, Median from 0.531m~(SDRDIS) to 0.378m in Site4, and RMSE from 2.102m~(S2P) to 1.993m in Site6.
% Compare to traditional methods, our TS-Sat-MVSNet exhibits SOTA performance on the vast majority of sites, e.g., Site2, $<$1.0m: 71.02\%~(Ours): 70.4\%~(S2P), Site4, Median: 0.378m~(Ours): 0.531~(SDRDIS), Site6, RMSE: 1.993m: 2.102m. 
% Moreover, compared to deep learning-based methods, our method achieves SOTA or comparable performance on the vast majority of sites, e.g., Site1, $<$1.0m: 73.9\%~(Ours): 68.47\%~(Sat-MVSF), Site3, Median: 0.321m~(Ours) vs. 0.338m~(Sat-MVSF), Site6, RMSE: 1.993m~(Ours): 2.086m~(Sat-MVSF). 
Moreover, compared to deep learning-based methods, our method also achieves SOTA or comparable performance on the vast majority of sites. For instance, our method can improve the $<$1.0m from 68.47\%~(Sat-MVSF) to 73.9\% in Site1, Median from 0.338m~(Sat-MVSF) to 0.321m in Site3, and RMSE from 2.086m~(Sat-MVSF) to 1.993m in Site6.
Although our method achieved SOTA performance on most sites and average metrics, there are still some sites where certain metrics are weaker than those of the methods compared above. We posit that the observed variance in model performance across different sites, particularly where our method underperforms in comparison to others, can be attributed to the diverse nature of the MVS3D dataset. This dataset spans a broad spectrum of scenes, encompassing urban landscapes, architectural wonders, and natural environments. Specifically, the complexity and heterogeneity inherent in urban areas may challenge the efficacy of our slope-guided manner, thereby impacting the overall performance of our model in such contexts.

\begin{figure*}[!t]
    \centering 
    \includegraphics[width=1\linewidth]{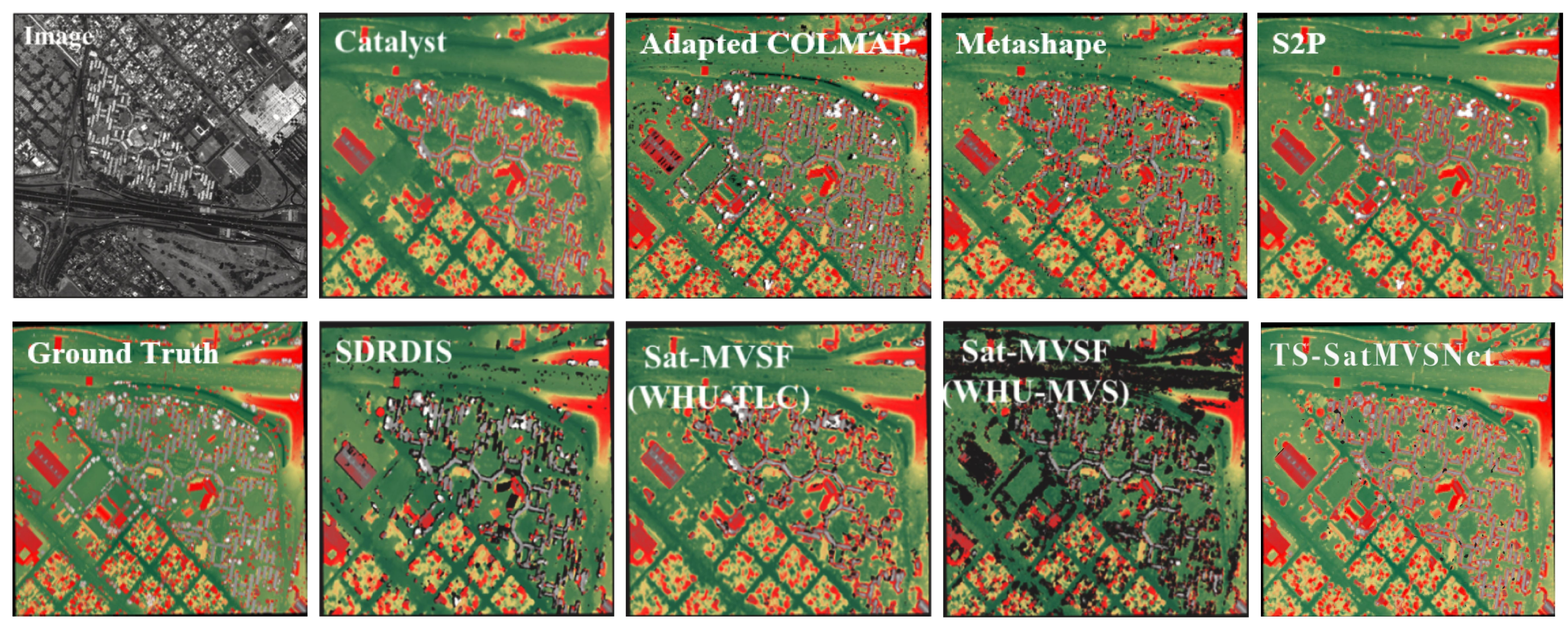}\vspace{-3mm}
    \caption{Visualization of the results of DSM evaluation in the first scene of the MVS3D dataset. Some qualitative results are obtained from Sat-MVSF.}
    \label{fig:figure_9} 
\end{figure*}

\input{mvs3d}

Furthermore, we also illustrate the qualitative results of DEM evaluation on the MVS3D testing set. As shown in Figure \ref{fig:figure_9}, we can observe that compared to other methods, our reconstructed DSM results do not contain much noise, i.e., there are not many white or black noise points ~(both black and white points in the image are considered invalid outliers in the DSM). We attribute this improvement in performance to our model's accurate height estimation, which enables the synthesis of DSM models with fewer noise points and higher precision.
In addition, it is worth mentioning that although our proposed method is not designed for height estimation in urban areas, we can still observe from Figure \ref{fig:figure_9} that the DSM results obtained by our method are comparable. This observation indicates that our method maintains the generalizability for urban areas where these areas are unsuitable for slope-guided manner.

% 对比一下在MVS3D上的结果
\begin{figure*}[!t]
    \centering 
    \includegraphics[width=1\linewidth]{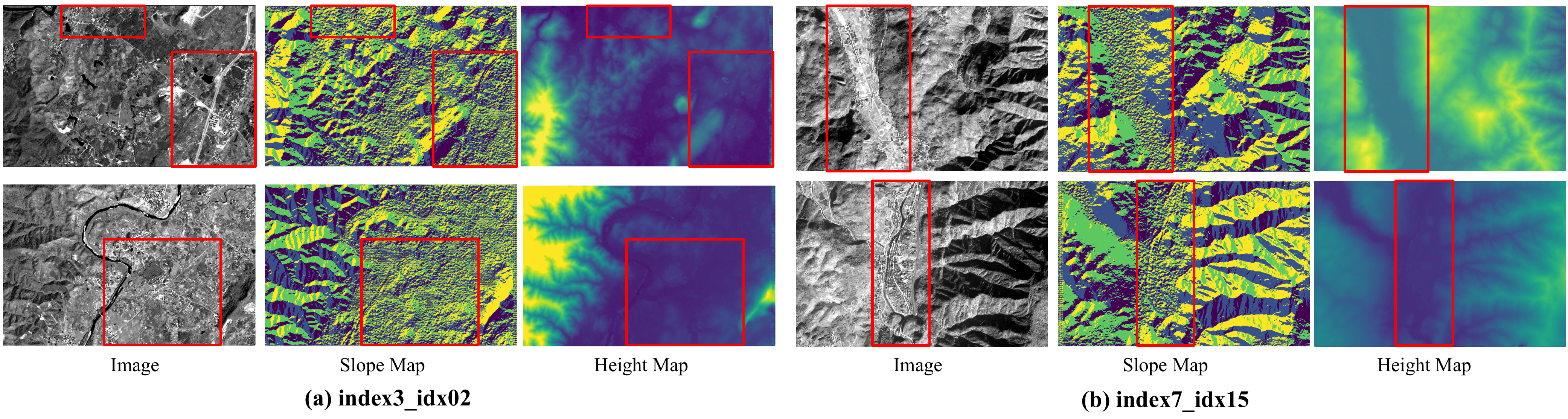}
    \vspace{-8mm}
    \caption{Visualization of some limited cases of mixed terrain and urban scenarios.}
    \label{fig:figure_10} 
    \vspace{-5mm}
\end{figure*}

\section{Limitation and Consideration}
% 主要局限是针对城市区域可能出现bad case（给几张图），但是也可以理解因为我们这里采用的是针对地形起伏来指定的策略。
In this study, a slope-aware height estimation pipeline has been leveraged for large-scale earth terrain scenarios, demonstrating significant effectiveness in extensive RS data. While our framework has successfully improved higher accuracy across a variety of benchmarks, it is imperative to acknowledge certain limitations that warrant further investigation: 
~\emph{Inapplicability on Urban Area:} Given the significant and abrupt elevation changes encountered within the urban region, the height based slope calculation strategy we have proposed may not be entirely applicable to this specific area. This misalignment could manifest in outcomes artifacts to those illustrated within the red boxes of the slope map presented in Figure \ref{fig:figure_10}. Therefore, the integration of slope information into the pipeline could potentially exacerbate the challenges faced by the model in accurately height estimating for this district. 
Future studies might focus on refining the height estimation for urban regions by designing a specialized module dedicated to accurately capturing the grid structural characteristics inherent to urban architecture.

% 由于城市区域高度突变十分剧烈，因此我们提出的坡度计算准则可能并不适用于该区域，从而导致出现如Figure 10中Slope Map红框区域所显示的结果，对于pipeline在融合slope信息后可能会进一步影响模型对于该地区高度估计的结果。
\section{Conclusion}
In conclusion, this paper presents TS-SatMVSNet, a novel slope-aware height estimation framework for large-scale earth terrain reconstruction, which capitalizes on the incorporating slope information to capture the undulations of the terrain to improve the height estimation results. Different from previous methods lacks considering terrain characteristics and leads to low accuracy, TS-SatMVSNet adopts an innovative approach by a height-based slope calculation strategy to calculate a slope map from a height map to further incorporate the terrain characteristics. Specifically, we separately designed a slope-guided interval partition module and a height correction module to achieve more accurate height estimation. Moreover, the overall framework is constrained by two individual losses, e.g., depth loss, slope direction loss. To ascertain the efficacy of TS-SatMVSNet, we meticulously carried out our experiments on two different datasets, e.g., WHU-TLC dataset and MVS3D dataset. Comprehensive experiments conducted on these datasets have underscored TS-SatMVSNet’s considerable impact on terrain height estimation task. Furthermore, through rigorous ablation studies, the crucial role of height-based slope calculation strategy and the incremental benefits of achieving the slope-guided manner have been affirmed, attesting to the robustness and indispensability of each component within our framework. Our future work aims to explore the broader implications of our approach across different domains while striving to refine the precision of height estimation techniques further.

\bibliography{aaai24}
\clearpage

\end{document}

%% file: tlc_mvs_dataset.tex
\begin{table*}[!t]
    \centering
    \scriptsize
    \resizebox{1.8\columnwidth}{!}{
        % \tablestyle{10pt}{1.0}
        \begin{tabular}{c|c|ccccc}
        \toprule[1.2pt]
            Methods & Image Size & MAE~(m) $\downarrow$ & RMSE~(m) $\downarrow$ & $<$2.5m~(\%) $\uparrow$ & $<$7.5m~(\%) $\uparrow$ & Comp~(\%) $\uparrow$ \\
            \hline
            %Adapted COLMAP & 2048 × 1472 & 2.227 & 5.291 & 73.35 & 96.00 & 79.10 \\
            %\hline
            CasMVSNet$^{*}$ & 2048 × 1472 & 2.031 & 4.351 & 77.39 & 96.53 & 82.33 \\
            SatMVS(CasMVSNet) & 2048 × 1472 & 2.020 & 3.841 & 76.79 & 96.73 & 81.54 \\
            UCS-Net$^{*}$ & 2048 × 1472 & 2.039 & 4.084 & 76.40 & 96.66 & 82.08 \\
            SatMVS(UCS-Net) & 2048 × 1472 & 2.026 & 3.921 & 77.01 & 96.54 & 82.21 \\
            SatMVS-F & 2048 × 1472 & 1.895 & \textbf{3.654} & 64.82 & 80.05 & - \\
            \hline
            \multirow{3}{*}{RED-Net$^{*}$} & 2048 × 1472 & 2.171 & 4.514 & 74.13 & 95.91 & 81.82 \\
             & 5120 × 5120 & 2.517 & 4.873 & 66.42 & 95.53 & 81.44 \\
            & & $\textcolor{blue}{\textbf{+15.9\%}}$  & $\textcolor{blue}{\textbf{+7.9\%}}$ & $\textcolor{blue}{\textbf{-10.4\%}}$ & $\textcolor{blue}{\textbf{-0.3\%}}$ & $\textcolor{blue}{\textbf{-0.4\%}}$ \\
            \hline
            \multirow{3}{*}{SatMVS(RED-Net)} & 2048 × 1472 & 1.945 & 4.071 & \textbf{77.93} & 96.59 & 82.29 \\
             & 5120 × 5120 & 1.946 & 4.224 & 77.88 & 96.54 & 82.35 \\
            & & $\textcolor{blue}{\textbf{+0.1\%}}$  & $\textcolor{blue}{\textbf{+3.8\%}}$ & $\textcolor{blue}{\textbf{-0.1\%}}$ & $\textcolor{blue}{\textbf{-0.1\%}}$ & $\textcolor{blue}{\textbf{-0.1\%}}$ \\
            \hline
            \multirow{3}{*}{TS-SatMVSNet}  & 2048 × 1472 & \textbf{1.879} & 3.892 & 77.92 & \textbf{97.34} & \textbf{82.60} \\
              & 5120 × 5120 & 1.882 & 3.919 & 77.90 & 97.32 & 82.52 \\
            & & $\textcolor{blue}{\textbf{+0.1\%}}$  & $\textcolor{blue}{\textbf{+0.6\%}}$ & $\textcolor{blue}{\textbf{-0.1\%}}$ & $\textcolor{blue}{\textbf{-0.1\%}}$ & $\textcolor{blue}{\textbf{-0.1\%}}$ \\
        \bottomrule
        \end{tabular}
        }
    \caption{The quantitative results on WHU-TLC Dataset \cite{satmvs} compared with the deep learning-based MVS methods. $^{*}$ represents the methods adopt pinhole. `-' represents the default values. `blue font'  indicates the percentage of performance improvement from a resolution of 5120×5120 to 2048×1472. Some results are obtained from SatMVS \cite{satmvs}.}
    \label{tab:Table.1}
    \vspace{-3mm}
\end{table*}

%% file: tlc_mvs_dataset_1.tex
\begin{table}[!t]
    \centering
    \scriptsize
    \resizebox{1\columnwidth}{!}{
        % \tablestyle{10pt}{1.0}
        \begin{tabular}{c|ccccc}
        \toprule[1.2pt]
            Methods & MAE~(m) $\downarrow$ & RMSE~(m) $\downarrow$ & $<$2.5m~(\%) $\uparrow$ & $<$7.5m~(\%) $\uparrow$ \\
            \hline
            S2P & 3.158 & 10.089 & 54.96 & 73.37 \\
            SDRDIS & 4.496 & 15.012 & 47.58 & 73.57 \\
            Adapted COLMAP & 2.168 & 4.714 & 58.78 & 76.80 \\
            ArcGIS & 4.607 & 10.689 & 48.88 & 77.71 \\
            CATALYST & 3.454 & 7.939 & 52.31 & 82.52 \\
            Metashape & 2.693 & 13.047 & 56.59 & 75.46 \\
            \hline
            TS-SatMVSNet & \textbf{1.879} & \textbf{3.892} & \textbf{77.92} & \textbf{97.34} \\
        \bottomrule
        \end{tabular}
        }
    \caption{The quantitative results on WHU-TLC Dataset \cite{satmvs}. $^{*}$ represents the methods adopt pinhole. Some results are obtained from SatMVSF \cite{satmvsf}. Some results are obtained from Sat-MVSF.}
    \vspace{-2mm}
    \label{tab:Table.2}
\end{table}

%% file: ablation_study.tex
\begin{table*}[!t]
    \centering
    \scriptsize
    \resizebox{0.85\linewidth}{!}{
        % \tablestyle{10pt}{1.0}
        \begin{tabular}{c|ccc|ccccc}
        \toprule[1.2pt]
            Methods & SIPM & HCM & Dataset & MAE~(m) $\downarrow$ & RMSE~(m) $\downarrow$ & $<$2.5m~(\%) $\uparrow$ & $<$7.5m~(\%) $\uparrow$ \\
            \hline
            Baseline & & & WHU-TLC & 2.026 & 3.921 & 77.01 & 96.54  \\
            Baseline + SIPM & $\surd $ & & WHU-TLC & 1.911 & 3.898 & 77.83 & 97.15  \\
            Baseline + HCM & & $\surd $ & WHU-TLC & 2.002 & 3.914 & 77.25 & 96.90  \\
            \hline
            \multirow{3}{*}{TS-SatMVSNet}
                & $\surd $ & $\surd $ & WHU-TLC & 1.879 & 3.892 & 77.92 & 97.34 \\
                & $\surd $ & $\surd $ & $\text{WHU-TLC}^{*}$ & 1.793 & 3.732 & 79.03 & 97.88  \\
                & & & & $\textcolor{blue}{\textbf{+4.6\%}}$ & $\textcolor{blue}{\textbf{+4.1\%}}$ & $\textcolor{blue}{\textbf{+1.4\%}}$ & $\textcolor{blue}{\textbf{+0.5\%}}$ \\
        \bottomrule
        \end{tabular}
        }
    \caption{Ablation study on the WHU-TLC Dataset, which demonstrates the effectiveness of different modules of our method. `$\text{WHU-TLC}^{*}$' represents the reconstructed WHU-TLC, which contains pure terrain data. `SIPM' represents our proposed Slope-guided Interval Parition Module and `HCM' represents the Height Correction Module.}
    \vspace{-2mm}
    \label{tab:Table.3}
\end{table*}

%% file: mvs3d.tex
\begin{table*}[!t]
    \centering
    \scriptsize
    \resizebox{1\linewidth}{!}{
        % \tablestyle{10pt}{1.0}
        \begin{tabular}{cc|ccccccccc}
        \toprule[1.2pt]
            Sites & Metrics & CATALYST & Metashape & S2P & SDRDIS & \makecell{JHU \\ APL} & \makecell{Adapted \\ COLMAP} & \makecell{Sat-MVSF \\ (WHU-TLC)} & \makecell{Sat-MVSF \\ (WHU-MVS)} & TS-SatMVSNet \\
            \hline
            \multirow{3}{*}{Mean of all sites}
                & $<$1.0m~(\%) & 58.915 & 56.73 & 59.49 & 56.67 & 55.19 & 50.38 & 55.90 & 51.09 & \textbf{60.635} \\
                & Median~(m) & 0.767 & 0.495 & 0.400 & 0.503 & 0.883 & 0.371 & 0.587 & 0.368 & \textbf{0.353} \\
                & RMSE~(m) & 4.323 & 3.464 & 4.778 & 4.166 & 4.896 & 8.397 & 3.867 & 2.957 & \textbf{2.898} \\ \hline
            \multirow{3}{*}{Site1}
                & $<$1.0m~(\%) & 72.31 & 67.61 & \textbf{74.42} & 69.82 & 68.09 & 63.21 & 68.47 & 51.44 & 73.93 \\
                & Median~(m) & 0.353 & 0.279 & \textbf{0.235} & 0.304 & 0.511 & 0.261 & 0.34 & 0.281 & 0.253 \\
                & RMSE~(m) & 2.913 & 2.495 & \textbf{2.416} & 2.772 & 3.156 & 3.468 & 2.83 & 2.079 & 2.495 \\
            \multirow{3}{*}{Site2}
                & $<$1.0m~(\%) & 64.57 & 65.65 & 70.46 & 63.82 & 61.91 & 55.64 & 63.78 & 69.15 & \textbf{71.02} \\
                & Median~(m) & 0.548 & 0.397 & 0.348 & 0.529 & 0.655 & \textbf{0.264} & 0.571 & 0.354 & 0.341 \\
                & RMSE~(m) & 2.037 & 1.872 & 1.836 & 2.038 & 2.182 & 5.464 & 2.045 & 1.781 & \textbf{1.731} \\
            \multirow{3}{*}{Site3}
                & $<$1.0m~(\%) & \textbf{58.92} & 54.22 & 55.06 & 57.65 & 54.05 & 46.7 & 54.47 & 50.34 & 56.33 \\
                & Median~(m) & 0.665 & 0.506 & 0.377 & 0.395 & 0.827 & \textbf{0.321} & 0.489 & 0.338 & \textbf{0.321} \\
                & RMSE~(m) & 4.311 & 3.898 & 3.874 & 3.912 & 4.581 & 9.596 & 3.795 & \textbf{3.124} & 3.178 \\
            \multirow{3}{*}{Site4}
                & $<$1.0m~(\%) & \textbf{43.86} & 38.68 & 40.16 & 41.37 & 41.94 & 28.83 & 39.64 & 24.4 & 41.56 \\
                & Median~(m) & 1.466 & 0.722 & 0.533 & 0.531 & 1.527 & 0.599 & 0.698 & 0.389 & \textbf{0.378} \\
                & RMSE~(m) & 10.319 & 7.214 & 12.873 & 7.844 & 11.749 & 19.138 & 7.8 & 5.96 & \textbf{5.03} \\
            \multirow{3}{*}{Site5}
                & $<$1.0m~(\%) & 65.58 & 66.01 & 70.48 & 63.65 & 61.73 & 64.22 & 63.59 & 70.04 & \textbf{71.33} \\
                & Median~(m) & 0.549 & 0.413 & 0.377 & 0.55 & 0.662 & \textbf{0.35} & 0.609 & 0.374 & 0.377 \\
                & RMSE~(m) & 2.127 & 1.821 & \textbf{1.772} & 2.055 & 2.24 & 2.777 & 2.142 & \textbf{1.772} & 1.990 \\
            \multirow{3}{*}{Site6}
                & $<$1.0m~(\%) & 63.32 & 62.85 & \textbf{67.47} & 61.17 & 57.44 & 60.88 & 60.65 & 66.23 & 67.03 \\
                & Median~(m) & 0.58 & 0.446 & 0.397 & 0.573 & 0.744 & 0.375 & 0.66 & 0.395 & \textbf{0.370} \\
                & RMSE~(m) & 2.611 & 2.105 & 2.102 & 2.519 & 2.63 & 3.29 & 2.743 & 2.086 & \textbf{1.993} \\
            \multirow{3}{*}{Site7}
                & $<$1.0m~(\%) & 42.26 & \textbf{44.66} & 44.38 & 40.91 & 42.16 & 37.02 & 41.8 & 33.41 & 42.88 \\
                & Median~(m) & 1.355 & 0.75 & 0.556 & 0.767 & 1.308 & 0.483 & 0.846 & 0.451 & \textbf{0.441} \\
                & RMSE~(m) & 6.192 & 4.651 & 6.353 & 4.652 & 8.162 & 13.192 & 5.869 & 3.265 & \textbf{3.096} \\
            \multirow{3}{*}{Site8}
                & $<$1.0m~(\%) & 60.50 & 54.14 & 53.50 & 54.95 & 54.19 & 46.51 & 54.82 & 43.71 & \textbf{61.00} \\
                & Median~(m) & 0.621 & 0.443 & 0.375 & 0.374 & 0.83 & \textbf{0.316} & 0.48 & 0.362 & 0.339 \\
                & RMSE~(m) & 4.077 & 3.657 & 6.996 & 7.539 & 4.467 & 10.254 & 3.713 & \textbf{3.592} & 3.667 \\
        \bottomrule
        \end{tabular}
        }
    \caption{Evaluation results obtained on the MVS3D dataset. The bold figures rank first and the underlined figures rank second in each metric for each site. }
    \label{tab:Table.4}
\end{table*}